\documentclass{article}

\PassOptionsToPackage{numbers, compress}{natbib}
\usepackage[preprint]{neurips_2023}

\usepackage{multirow}
\usepackage{adjustbox}
\usepackage{hyperref}
\usepackage{todonotes}

\usepackage{xcolor}

\begin{document}
%
\title{In-Domain Self-Supervised Learning Improves Remote Sensing Image Scene Classification}

\author{Ivica~Dimitrovski $^{1,2}$,
        Ivan~Kitanovski $^{1,2}$,
        Nikola~Simidjievski $^{1,3,4}$,
        Dragi~Kocev $^{1,3}$ \\ \\
        $^{1}$Bias Variance Labs, Slovenia\\
        $^{2}$Faculty of Computer Science and Engineering, University 'Ss. Cyril and Methodius', N. Macedonia\\
        $^{3}$Jozef Stefan Institute, Slovenia\\
        $^{4}$University of Cambridge, United Kingdom\\
        \texttt{[ivica,ivan,nikola,dragi]@bvlabs.ai}
}
\maketitle

\begin{abstract}
We investigate the utility of in-domain self-supervised pre-training of vision models in the analysis of remote sensing imagery. Self-supervised learning (SSL) has emerged as a promising approach for remote sensing image classification due to its ability to exploit large amounts of unlabeled data. Unlike traditional supervised learning, SSL aims to learn representations of data without the need for explicit labels. This is achieved by formulating auxiliary tasks that can be used for pre-training models before fine-tuning them on a given downstream task. A common approach in practice to SSL pre-training is utilizing standard pre-training datasets, such as ImageNet. While relevant, such a general approach can have a sub-optimal influence on the downstream performance of models, especially on tasks from challenging domains such as remote sensing. In this paper, we analyze the effectiveness of SSL pre-training by employing the iBOT framework coupled with Vision transformers trained on Million-AID, a large and unlabeled remote sensing dataset. We present a comprehensive study of different self-supervised pre-training strategies and evaluate their effect across 14 downstream datasets with diverse properties. Our results demonstrate that leveraging large in-domain datasets for self-supervised pre-training consistently leads to improved predictive downstream performance, compared to the standard approaches found in practice. \looseness=-1

\end{abstract}

\textbf{Keywords:} Remote Sensing, Self-Supervised Learning, Earth Observation, Deep Learning

\section{Introduction}\label{sec:intro}

The growth in the availability of remote sensing data is matched only by the growth and developments in the field of artificial intelligence (AI) and, in particular, computer vision. Recent trends in deep learning have led to a new era of image analysis and have raised the bar for predictive performance in many application areas, including remote sensing and Earth observation (EO). Deep learning models can leverage these large datasets to achieve state-of-the-art performance on various EO tasks (including land use/land cover classification, and crop prediction) using large amounts of labeled remote sensing data (typically hundreds of thousands of labeled images).

However, for many highly relevant tasks (e.g., archaeological site identification, pasture grazing, agricultural fertilization), there is a lack of (sufficiently) large, publicly available, and more importantly - labeled data. This can be a limiting factor for further uptake and use of recent AI advancements for EO tasks. On the one hand, this is understandable and to be expected -- given that annotation of large datasets is typically an expensive, tedious, time-consuming, and largely manual process. On the other hand, such scenarios call for methods that can learn and represent the (visual) information contained in images without the need for labeled examples. Such methods build on the self-supervised learning (SSL) paradigm \cite{WangYi2022}. \looseness=-1

SSL provides a novel approach to the data labeling challenge by leveraging the benefits of available (unlabeled) data without the additional overhead of manual labeling. Within the SSL paradigm, a model is trained in two stages, an upstream and a downstream stage. In the upstream stage, a model is first pre-trained on a large unlabeled dataset. In the downstream stage, the self-supervised pre-trained model is further fine-tuned using fully supervised training and applied to a specific downstream task with a limited amount of labeled data. In the context of remote sensing, the downstream tasks also referred to as target tasks, can include scene classification, semantic segmentation, representation learning, and object detection \cite{WangYi2022}.

The main idea behind SSL, is to utilize an auxiliary, pretext, task in the upstream stage, directly from the input data. While learning to solve the pretext task, the model learns useful and relevant representations and features of the input data and its underlying structure, which is beneficial for the downstream tasks. There are several types of pretext tasks, such as augmenting the input data and then training a model to learn the applied augmentation, e.g., rotation, color removal, repositioning/removing patches, etc. in the images. 

In the context of remote sensing applications, there is a recent increased interest in studying and leveraging SSL approaches. Related work in this area partly focuses on introducing new datasets that can be used for (pre-)training such as SSL4EO-S12 \cite{wang2022ssl4eo} and Satlas \cite{bastani2022satlas}. Here the SSL approaches\cite{caron2021emerging,Kaiming2022} are typically evaluated through the prism of the pre-training dataset and its benefits to the models' overall performance. Other related studies focus on learning paradigms or model architectures, such as employing multi-augmentation contrastive learning \cite{Manas2021} to account for seasonal changes; novel multi-modal contrastive learning approach \cite{10147273} or novel attention mechanism \cite{wang_vitrvsa_2022}. \looseness=-1

In this work, we focus on the bigger picture and ask the fundamental question: \emph{Can in-domain self-supervised learning lead to consistent performance improvements on remote sensing tasks?} \looseness=-1

To assess this, we evaluate models on (downstream) tasks of image scene classification from 14 datasets. As a basis for SSL pre-training, we employ the iBOT framework \cite{zhou2021ibot} together with a Vision Transformer (ViT) \cite{dosovitskiy2020image} as the backbone model. iBOT (image BERT pre-training with Online Tokenizer) is a novel self-supervised pre-training framework that builds on masked image modeling combined with self-distillation. A recent comprehensive overview of SSL methods in remote sensing \cite{WangYi2022} reports that the iBOT framework (when applied with the ImageNet dataset) achieves state-of-the-art performance. Following these findings, we select iBOT as the base SSL framework in our work. Such an approach of pure in-domain pre-training has not been explored in the context of remote sensing applications. Another important aspect of the overall performance relates to the choice of a backbone model, which in the case of iBOT, is typically an attentional deep network architecture. In this work, we focus on Vision Transformers, a recent architecture with a track record of state-of-the-art performance on various vision tasks, especially in many remote sensing domains~\cite{dimitrovski2023}. 

In addition to our central question, in this work, we also aim to address several limitations of related studies that pertain to the scale, scope and reproducibility of the experimental workflows. Existing studies on SSL for remote sensing are often limited in scope, focusing on a small number of datasets, rarely exceeding three. Moreover, there is typically a lack of publicly available, reproducible, and reusable experimental workflows and details.
 Our study addresses these limitations by making the following contributions:
\begin{itemize}
    \item Scale: We conduct a \emph{comprehensive experimental study} of evaluating SSL pre-training strategies on downstream tasks from 14 remote sensing image scene classification datasets. We investigate the effect of pure in-domain pre-training and compare it to the standard approaches of utilizing ImageNet as a pre-training dataset;
    \item Scope: We evaluate these effects on \emph{diverse tasks}, from datasets that differ in several dimensions: Datasets with different numbers of images (from $\sim2K$ to $\sim500K$) at different spatial resolution, different labels per image (from $1$ up to $60$) and with varying label distribution (including balanced datasets as well as highly imbalanced datasets); 
    \item Reproducibility: We provide clearly defined protocols to standardize the experimental work and facilitate \emph{reproducibility and reusability} of the obtained models and results, including data pre-processing details, thus ensuring verifiable comparisons to other studies.\looseness=-1
\end{itemize}

In the reminder, we first elaborate on the iBOT self-supervised framework and introduce the remote sensing image scene classification as a downstream task, used to evaluate the performance of pre-trained strategies. Next, we outline the experimental design of the study, including the implementation details and settings. Finally, we present and discuss the findings, highlighting guidelines for further work.

\section{Methods and materials}

\begin{figure}[h]
    \centering
    \includegraphics[width=0.8\linewidth]{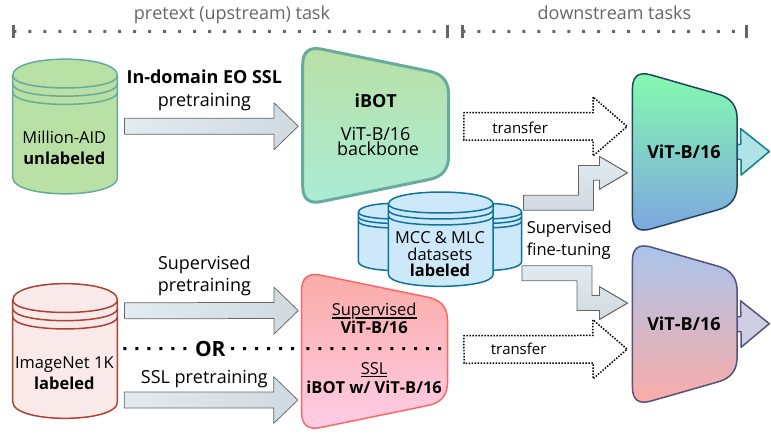}
    \caption{Illustration of the experimental pipeline. (Top) An \textit{An in-domain self-supervised pre-training strategy}: Using the self-supervised iBOT framework (with ViT-B/16 backbone) with unlabeled Million-AID dataset. (Bottom) A standard pre-training strategy with ImageNet 1K: Models (ViT-B/16) are either pre-trained in a fully supervised manner \textbf{or} using the iBOT framework (with ViT-B/16 backbone) for self-supervised pre-training on ImageNet. All models are then fine-tuned on each of the downstream datasets, separately. We evaluate and report the predictive performance of the resulting fine-tuned models.\looseness=-1}
    \label{fig:diagram}
\end{figure}

\subsection{Self-supervised pre-training with iBOT}

The iBOT ({\textbf i}mage {\textbf B}ERT pre-training with {\textbf O}nline {\textbf T}okenizer) framework \cite{zhou2021ibot} builds on the concept of Masked Image Modeling (MIM), an extension of Masked Language Models (MLM) for vision tasks. Much like MLM, in which a set of input tokens is randomly masked and reconstructed, MIM revolves around the design of a visual tokenizer, responsible for converting masked image patches into supervisory signals. \looseness=-1 

This approach, referred to as self-distillation, employs coupled models, where one serves as a “teacher” and the other as a “student” for learning the upstream pretext tasks. The former (online tokenizer) produces visual tokens of patches (or pixels) from the original, unaltered image. The latter (target model) attempts to learn the pretext task independently, by recovering it from the augmented/masked token and matching it against the corresponding output produced by the online tokenizer. As such, iBOT provides an effective and improved training pipeline for transformer architectures, such as the Vision Transformer. Vision Transformers~\cite{dosovitskiy2020image} are a class of recent attentional deep network architectures. They are inspired by the popular Transformer architectures for natural language processing, leveraging an attention mechanism for vision tasks. Much like the original Transformers that seek to learn implicit relationships in sequences of word-tokens via multi-head self-attention, ViTs focus on learning such relationships between image patches. The transformers have shown excellent performance on various vision tasks, especially in many remote sensing domains \cite{dimitrovski2023}, particularly when combined with pre-training from large datasets. \looseness=-1

\subsection{Unlabeled dataset for self-supervised pre-training}

At the core of defining a self-supervised pretext task is the (unlabeled) data, utilized for learning good-quality image representations. In our study, we employ the Million Aerial Image Dataset (Million-AID) \cite{long2021creating} dataset with remote sensing scene images. It aligns well with the desirable properties of many SSL pre-training datasets -- diversity, richness, quality, and size. \looseness=-1

The Million-AID dataset comprises 1,000,848 non-overlapping remote sensing scenes obtained from Google Earth, hierarchically organized into 95 categories. The images are in RGB format, which is a typical format used as input to deep learning models, especially in the context of remote sensing image scene classification. The size of images in the dataset varies from 110$\times$110 to 31,672$\times$31,672 pixels, due to the fact that they were captured using different types of sensors. In the pre-training phase, we use the Million-AID images without any labels/queries used for its construction. As such, the resulting pre-trained models can then be applied to different downstream tasks.

\subsection{Downstream tasks}

Evaluating the performance of SSL methods involves assessing their effectiveness on the downstream/target tasks. In particular, the resulting pre-trained models, obtained in the self-supervised learning stage are transferred to these specific downstream tasks. Transfer learning facilitates the use of learned representations from models pre-trained on significantly larger image datasets. This enables downstream models to benefit from such knowledge, often resulting in improved generalization power while requiring less training data and fewer iterations for fine-tuning. This advantage is particularly valuable for tasks that underline smaller datasets.

In this work, we focus on image scene classification, as downstream tasks. In a typical remote sensing scenario working with large-scale imagery, the task is to classify smaller images (patches), usually extracted from a much larger remote sensing image. The extracted images can then be annotated with one or more labels based on the content using explicit semantic classes (e.g., forests, residential areas, rivers, etc.). Given an image as input, the model must output a single or multiple semantic labels that denote the land use and/or land cover classes present in that image.

Based on the number of semantic labels assigned to the images in the datasets, image scene classification tasks can be categorized into multi-class (MCC) and multi-label (MLC) classification. In MCC, each image is associated with a single class or label from a predefined set, and the goal is to predict exactly one class for each image in the dataset. Conversely, in MLC, images can be associated with multiple labels from a predefined set, and the goal is to predict the complete set of labels for each image in the dataset.

We use 9 MCC datasets and 5 MLC datasets to benchmark and evaluate the performance of the self-supervised models in different contexts. The MCC datasets include Eurosat, UC Merced, Aerial Image Dataset (AID), RSSCN7, Siri-Whu, Resisc45, CLRS, RSD46-WHU, and Optimal31. 
The MLC datasets include AID (mlc), UC Merced (mlc), MLRSNet, Planet UAS, and BigEarthNet (the version with 19 labels). A more detailed description of these datasets is given in \cite{dimitrovski2023}.

\section{Experimental setup}

We evaluate our main research hypothesis (posed in Section~\ref{sec:intro}) by comparing the downstream effects of using either an in-domain or standard out-domain (ImageNet) pre-training dataset. Specifically, we evaluate 3 different pre-training scenarios: (1) models pre-trained with self-supervised iBOT using the unlabeled Million-AID data; (2) models pre-trained with self-supervised iBOT using standard unlabeled ImageNet-1k data (3) models that have been pre-trained in a fully supervised manner using labeled ImageNet-1k data. Figure~\ref{fig:diagram} illustrates the experimental pipelines for pre-training. All resulting pre-trained models are then fine-tuned on each of the downstream datasets, separately. We also provide a comparison to models that have been trained 'from scratch' on the downstream dataset, without any pre-training.

\subsection{Transfer learning strategies}

We employ two strategies for transfer learning, i.e. applying the learned pre-trained models to the downstream tasks: (1) Linear probing (or linear classification) - by updating the model weights only for the last, classifier, layer or (2) Fine-tuning the model weights of all layers in the network. The former approach retains the values of all but the last layer's weights of the model from the pre-training, keeping them 'frozen' during the training on the downstream task. The latter, on the other hand, allows the weights to change throughout the entire network during fine-tuning on the downstream task. Note that, for the models trained from scratch, the models are initialized with random weights.

\subsection{Implementation details and settings}

\begin{table*}[!t]

\caption{Performance on MCC datasets in terms of accuracy (top) and MLC datasets in terms of mean average precision (bottom). Results include models trained from scratch, pre-trained using ImageNet (both supervised and self-supervised), and self-supervised pre-trained using Million-AID (EO SSL). Italic indicates the best performance with Linear Probing, while bold indicates the best overall performance. In-domain SSL pre-training, consistently leads to improved performance, across all tasks.}
\label{tbl:results}
\centering

\begin{adjustbox}{width=1\textwidth,center}
\begin{tabular}{|l|c|ccc|ccc|}
\hline

\multirow{3}[0]{*}{Dataset} & \multirow{3}[0]{*}{\shortstack{Training \\ from scratch}}
& \multicolumn{3}{c|}{Linear Probing} & \multicolumn{3}{c|}{Fine-Tuning} \\

      & 
      & ImageNet & ImageNet & In-Domain & ImageNet & ImageNet & In-Domain \\
      
      & & Supervised & SSL & EO SSL & Supervised & SSL &  EO SSL \\
\hline
\hline
Eurosat  & 95.037 & 97.056 & 96.389 & \textit{97.981} & 98.722 & 98.759 &  \textbf{98.963} \\
UC merced  & 83.095 & 93.333 & 95.952 & \textit{96.190} & 98.333 & 97.381 &  \textbf{98.810} \\
AID  & 79.350 & 94.950 & 95.150 & \textit{97.550} & 97.750 & 97.200 &  \textbf{97.800} \\
RSSCN7  & 86.071 & 89.464 & 90.000 & \textit{92.321} & 95.893 & 95.000 &  \textbf{96.607} \\
SIRI-WHU  & 86.250 & 89.792 & 90.625 & \textit{93.542} & 95.625 & 96.250 &  \textbf{97.708} \\
RESISC45  & 81.016 & 92.016 & 91.698 & \textit{95.651} & 97.079 & 95.873 &  \textbf{97.254} \\
CLRS  & 65.467 & 87.200 & 87.233 & \textit{91.133} & 93.200 & 91.300 &  \textbf{93.367} \\
RSD46-WHU  & 86.466 & 89.510 & 89.230 & \textit{92.051} & 94.238 & 93.062 &  \textbf{94.518} \\
Optimal31  & 62.634 & 83.602 & 83.333 & \textit{91.129} & 94.624 & 88.978 &  \textbf{95.699} \\
\hline
\hline
AID (mlc)  & 65.581 & 74.094 & 72.446 & \textit{78.096} & 81.540 & 79.589 &  \textbf{82.543} \\
UC merced (mlc)  & 87.142 & 89.746 & 90.210 &\textit{ 90.320 }& 96.699 & 96.363 &  \textbf{97.053} \\ 
MLRSNet  & 87.250 & 90.083 & 89.313 & \textit{94.225} & 96.410 & 95.558 &  \textbf{96.837 }\\
Planet UAS  & 59.414 & 62.710 & 62.791 & \textit{64.059} & 66.804 & 65.832 &  \textbf{67.139} \\
BigEarthNet  & 75.871 & 68.357 & 67.007 & 70.961 & 77.310 & 79.198 &  \textbf{79.361} \\
\hline
\end{tabular}
\end{adjustbox}
\end{table*}

For the downstream tasks, we use the train, validation, and test splits of the image datasets (60\%, 20\%, and 20\% fractions, respectively) as provided in \cite{dimitrovski2023}. We use the train splits for model training, while the validation splits for early stopping of the training process - thus preventing overfitting. Models with the lowest validation loss are saved and evaluated on the test split to estimate their predictive performance. In all experiments, we use the ViT-Base model with an input size of 224$\times$224 and a patch resolution of 16$\times$16 pixels (ViT-B/16). For in-domain SSL pre-training, we train iBOT with ViT-B/16 as the backbone for 400 epochs using the Million-AID dataset. For the ImageNet SSL pre-training variant, we use the official model weights from \cite{zhou2021ibot}, which has also been trained for 400 epochs. Finally, the fully-supervised pre-trained variant utilizes pre-trained ViT-B/16, available from the \texttt{timm} repository \cite{rw2019timm}.

To further enhance the model's robustness, we incorporate \emph{data augmentation} techniques during training. The process involves resizing all images to 256$\times$256 pixels, followed by random crops of size 224$\times$224 pixels. Additionally, random horizontal and/or vertical flips are applied. During the evaluation of predictive performance, when applying the model to test data, the images are first resized to 256$\times$256 pixels and then subjected to a central crop of size 224$\times$224 pixels. We believe that these augmentation techniques contribute to better generalization of our models for a given dataset. 

During training, we follow the parameter settings suggested in the literature (cf. \cite{zhou2021ibot}). We evaluate the performance of the models learned using different values of learning rate: $0.01-0.00001$. Next, we use \emph{ReduceLROnPlateau} as a learning scheduler, reducing the learning rate when the loss has stopped improving. Models often benefit from reducing the learning rate once learning stagnates for a certain number of epochs (denoted as `patience'). In our experiments, we track the value of the validation loss, with patience set to $5$ and reduction factor set to $0.1$. The maximum number of epochs for fine-tuning is set to $100$. We also apply early stop criteria if no improvements in the validation loss are observed over 10 epochs. We use fixed values for batch size, set to $128$. For optimization, we employ \emph{RAdam optimizer} without weight decay \cite{liu2019radam}.\looseness=-1 

In terms of evaluation measures for predictive performance, we report 'top-1 accuracy' measure (referred to as 'Accuracy') in the case of multi-class classification tasks, and macro-averaged 'mean-average precision' (mAP) for multi-label classification tasks.\looseness=-1

All models were trained on NVIDIA A100-PCIe GPUs with 40 GB of memory running CUDA version 11.5. We configured and ran the experiments using the AiTLAS toolbox \cite{dimitrovski2023:aitlas} (available at \url{https:\\aitlas.bvlabs.ai}). All configuration files, for each experiment, as well as the trained model weights, can be found at \url{https://aitlas-arena.bvlabs.ai}. \looseness=-1

\section{Results}

Table \ref{tbl:results} presents the results of our study. More precisely, it showcases the performance of models trained with different configurations, comparing the effect of different pre-training strategies as well as the different transfer learning strategies on the downstream model performance across all 14 downstream tasks. This includes models that utilize in-domain SSL pre-training denoted as 'In-domain EO SSL'; models that have been pre-trained using ImageNet, either in a fully-supervised or in an SSL setting (denoted as 'ImageNet Supervised' and 'ImageNet SSL', respectively); and models trained from scratch, directly on the downstream dataset. Note that, based on the type of transfer mechanism, we also compare two different variants of models that leverage pre-training, which relate to models using linear probing and models that have been fine-tuned. Such a comparison allows us to systematically examine various factors in the learning process, and better understand their effect on final model performance. 

The main conclusion from these analyses is that employing \textit{in-domain} SSL pre-training, consistently leads to improved performance, across all downstream tasks. Using in-domain pre-training leads to better performance compared to models that have been pre-trained from ImageNet, either in a fully supervised or self-supervised manner, as it is commonly done in practice. When compared to models that have been pre-trained in a supervised manner, in-domain SSL pre-training leads to $\sim1\%$ across tasks - which, while not substantial, is consistent. The improvements are larger when compared to models that have been SSL pre-trained from ImageNet. In this case, in-domain SSL pre-training leads to up to 2\% when compared to their counterparts. These results point to the fact that, while SSL pre-training is very valuable, the choice of the pre-training dataset also plays a crucial role in the downstream model performance. As such, this directly confirms our central question: \textit{In-domain self-supervised learning leads to consistent performance improvements on remote sensing tasks}.

The benefits of in-domain pre-training also extend to both transfer learning strategies. In general, pre-trained models that have been fine-tuned lead to much larger improvements than models that employ linear probing. In particular, when compared to models trained from scratch, fine-tuned models lead to substantial improvements, on average $\sim$16.15\% in the case of MCC and $\sim$9.54\% in the case of MLC downstream tasks. These performance improvements are mostly consistent across different pre-training strategies, with in-domain SSL pre-training outperforming both ImageNet counterparts. A notable exception is the case of BigEarthNet, where models trained from scratch outperform pre-trained models with linear probing. Here we believe this is due to the particular dataset properties, and as such this behavior is somewhat expected. BigEarthNet contains a large amount of good-quality images (500K images in total), sufficient to train a model from scratch with practically good performance. Therefore, the benefits of combining pre-training with linear probing are undermined, and as evident, fine-tuning should be the preferred transfer learning strategy. 

\begin{figure}
    \centering
    \includegraphics[width=0.5\linewidth]{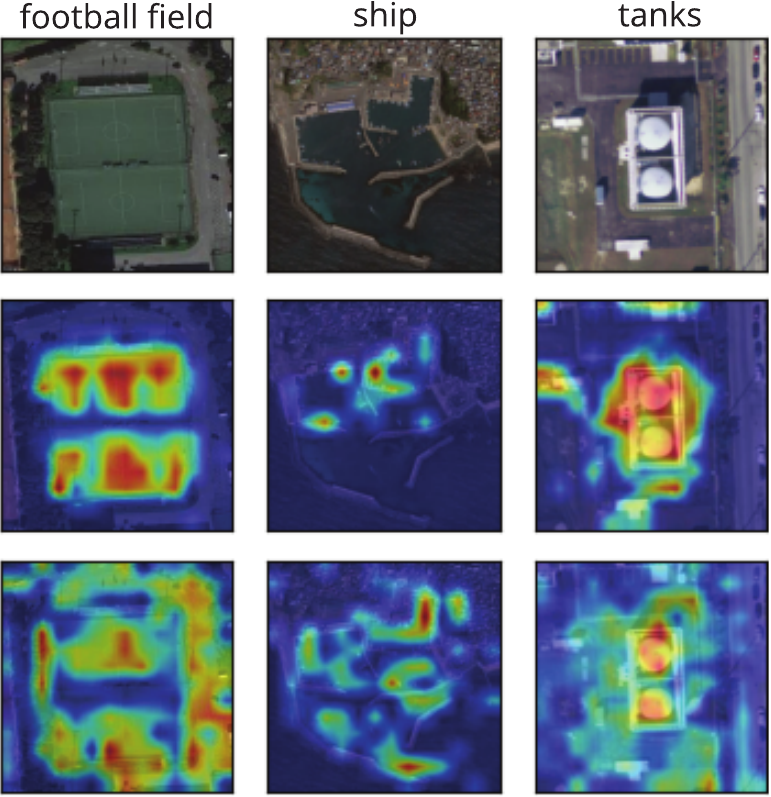}
    \caption{GradCAM visualizations for the predictions made with fine-tuned models on images sampled from the MLC datasets (`football field' from MLRSNet, `ship' from AID mlc, and `tanks' from UC Merced mlc): (\textit{first row}) Input images with their ground-truth label, (\textit{second row}) corresponding activation maps from the SSL pre-trained model, (\textit{third row}) corresponding activation maps from the supervised pre-trained model.}
    \label{fig:gradcam_examples}
\end{figure}

Finally, we performed further analysis of model performance, in terms of investigating their accuracy across different labels from the individual downstream datasets. We found that, in general, the largest differences in performance between models pre-trained with either Million-AID or ImageNet are often observed in cases that are scarcely annotated i.e. labels that annotate only a limited number of images. Such cases are typically present in MLC datasets. To showcase these observations, we present activation maps using GradCAM \cite{selvaraju2017} focusing on selected labels from the MLC datasets (Figure~\ref{fig:gradcam_examples}). These maps show that the models that have been in-domain SSL pre-trained correctly identify regions relevant to the true label assigned to the image. The identified regions are more focused on the specific objects of interest as compared to the identified regions by the models that have been fully-supervised pre-trained using ImageNet. This suggests that in-domain pre-training has a stronger capability to discern and highlight the specific objects associated with the labels, leading to improved performance in challenging such scenarios. \looseness=-1

\section{Conclusion}

We present a comprehensive study on the use of in-domain self-supervised pre-training of models in the analysis of remote sensing imagery. The presented analysis provides strong evidence that leveraging large in-domain datasets for self-supervised pre-training leads to improved predictive downstream performance, compared to the standard approaches found in practice of using ImageNet in a fully supervised or self-supervised manner. Our analysis is supported by experiments from 14 datasets, diverse in the number of images and labels, their spatial resolution, label distribution, and downstream classification tasks. 

Our findings demonstrate that employing \textit{in-domain SSL pre-training leads to better performance} than common pre-training strategies that rely on out-domain datasets, such as ImageNet. The performance improvements are consistent across all considered downstream tasks, regardless of the transfer learning strategy. Moreover, compared to models trained from scratch, the performance improvements are also substantial, re-confirming that SSL pre-training - and in this case using in-domain data - is crucial for well-generalizable models. 

An immediate line of further work considers studying and understanding the performance of in-domain SSL models in relation to the different dataset properties as well as to the semantic meaning of the labels. This entails designing a variety of specialized ablation studies, generating artificial datasets, and exploring different frameworks for SSL model training.




\section*{Acknowledgment}
We acknowledge the support of the European Space Agency ESA through the activity AiTLAS - Artificial Intelligence toolbox for Earth Observation (ESA RFP/3-16371/19/I-NB) awarded to Bias Variance Labs, d.o.o..

\bibliographystyle{plain}

\bibliography{arxiv}



\end{document}